\documentclass[letterpaper,journal]{IEEEtran}
\usepackage{amsmath,amsfonts}
\usepackage{algorithm}
\usepackage{array}
\usepackage[caption=false,font=normalsize,labelfont=sf,textfont=sf]{subfig}
\usepackage{textcomp}
\usepackage{stfloats}
\usepackage{url}
\usepackage{verbatim}
\usepackage{algorithm}
\usepackage{algpseudocode}
\usepackage{graphicx}
\usepackage{cite}
\usepackage[table]{xcolor} 
\usepackage{multirow}      
\usepackage{booktabs}      
\usepackage{colortbl}         
\usepackage{pifont}
\usepackage{makecell} 
\usepackage{xcolor}
\definecolor{car}{RGB}{100,150,245}
\definecolor{bicycle}{RGB}{100,230,245}
\definecolor{motorcycle}{RGB}{30,60,150}
\definecolor{truck}{RGB}{80,30,180}
\definecolor{othervehicle}{RGB}{100,80,250}
\definecolor{person}{RGB}{255,30,30}
\definecolor{bicyclist}{RGB}{255,40,200}
\definecolor{motorcyclist}{RGB}{150,30,90}
\definecolor{road}{RGB}{255,0,255}
\definecolor{parking}{RGB}{255,150,255}
\definecolor{sidewalk}{RGB}{75,0,75}
\definecolor{otherground}{RGB}{175,0,75}
\definecolor{building}{RGB}{255,200,0}
\definecolor{fence}{RGB}{255,120,50}
\definecolor{vegetation}{RGB}{0,175,0}
\definecolor{trunk}{RGB}{135,60,0}
\definecolor{terrain}{RGB}{150,240,80}
\definecolor{pole}{RGB}{255,240,150}
\definecolor{trafficsign}{RGB}{255,0,0}

\usepackage{titlesec}
\titlespacing*{\section}{0pt}{4pt}{2pt}
\titlespacing*{\subsection}{0pt}{3pt}{1pt}
\begin{document}

\title{VOIC: Visible–Occluded Integrated Guidance for 3D Semantic Scene Completion}

\author{%
    Zaidao Han,
    Risa Higashita,
    Jiang Liu,~\IEEEmembership{Senior Member,~IEEE}%
    \thanks{This work was supported in part by the National Key R\&D Program of China (No.~2024YFE0198100) and the Shenzhen Medical Research Funds (Grant No.~D2402014).}%
    \thanks{Zaidao Han is with the Research Institute of Trustworthy Autonomous Systems, Southern University of Science and Technology, Shenzhen 518055, China, and also with the Department of Computer Science and Engineering, Southern University of Science and Technology, Shenzhen 518055, China (email: 12445017@mail.sustech.edu.cn).}%
    \thanks{Risa Higashita (Corresponding author) is with the Research Institute of Trustworthy Autonomous Systems, Southern University of Science and Technology, Shenzhen 518055, China, and also with the Department of Computer Science and Engineering, Southern University of Science and Technology, Shenzhen 518055, China (email: risa@mail.sustech.edu.cn).}%
    \thanks{Jiang Liu (Corresponding author) is with the Research Institute of Trustworthy Autonomous Systems, and the Department of Computer Science and Engineering, Southern University of Science and Technology, Shenzhen 518055, China; the School of Computer Science, University of Nottingham Ningbo China, Ningbo 315100, China; and the Department of Electronic and Information Engineering, Changchun University, Changchun 130022, China (email: liuj@sustech.edu.cn).}%
}

\maketitle
\IEEEpubidadjcol

\begin{abstract}

Camera-based 3D Semantic Scene Completion (SSC) aims to recover a complete 3D representation of both scene geometry and semantics from a single image, serving as a fundamental task in autonomous driving and robotic perception. Existing methods typically rely on uniform global supervision derived from full-scene annotations. However, such supervision ignores the intrinsic observability differences between visible and occluded regions, resulting in supervision contamination: unreliable occluded-region signals interfere with the learning of high-confidence visible regions, thereby degrading feature quality and limiting the effectiveness of global scene completion.
To address this issue, this work proposes an offline Visible Region Label Extraction (VRLE) strategy, which explicitly separates visibility-aware voxel-level supervision from dense 3D ground truth, providing clean and targeted learning objectives for visible-region perception. Building upon this, we introduce the Visible–Occluded Interactive Completion Network (VOIC), a dual-decoder architecture that formulates SSC as two complementary processes: visible-region semantic perception and holistic scene completion.
Specifically, VOIC first constructs a base 3D voxel representation by fusing image features with depth-derived occupancy. The Visible Decoder (VD) learns reliable geometric and semantic priors under explicit visibility-aware supervision. These priors serve as structured guidance for the Occlusion Decoder (OD) to infer the complete scene, while the OD simultaneously provides global contextual feedback to refine the visible-region features. This bidirectional interaction ensures spatial consistency and mutual optimization between observed and unobserved regions, improving the overall stability of the completion system.
Extensive experiments on the SemanticKITTI and SSCBench-KITTI-360 benchmarks demonstrate that the proposed method achieves competitive performance in both geometric completion and semantic segmentation. Code and pre-trained models are available at: https://github.com/dzrdzrdzr/VOIC.

\end{abstract}

\begin{IEEEkeywords}
Semantic Scene Completion, 
Visible-Occluded Decoupling, 
Autonomous Driving, 
Visible Region Label Extraction
\end{IEEEkeywords}

\section{Introduction}

\begin{figure}[!t]
  \centering
  \includegraphics[width=\columnwidth]{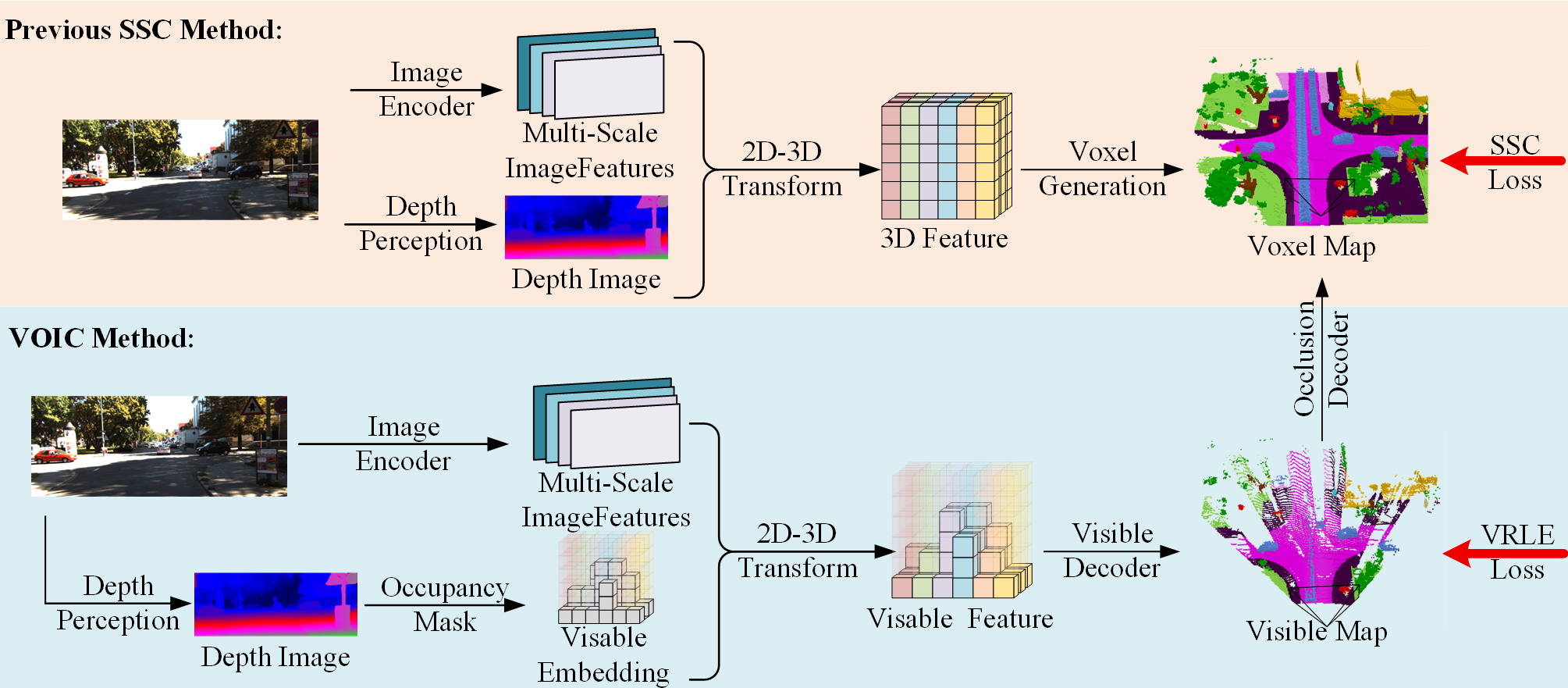}
  \caption{
  Overview of the proposed VOIC framework. 
Unlike conventional SSC methods that supervise with full 3D labels, {VOIC} employs a decoupled progressive pipeline with intermediate supervision. 2D features are first lifted to 3D via cross-modal fusion. The {Visible Decoder (VD)} handles observed regions, and the {Occlusion Decoder (OD)} leverages normalized VD features as priors to reconstruct the full 3D scene. VD and OD are supervised with {VRLE labels} and global labels, respectively, ensuring accurate and stable completion.
  }
  \label{overall-fig}

\end{figure}

3D Semantic Scene Completion (SSC) aims to infer the complete 3D geometry and semantic structure of a scene from partial visual observations, such as monocular or stereo images. As a fundamental task in scene understanding, SSC provides dense and structured 3D representations of the surrounding environment, serving as a crucial bridge between perception and reconstruction. Accurate SSC is critical for numerous downstream applications, including autonomous driving, robotic navigation, and mixed reality \cite{caoMonosceneMonocular3d2022a, chengS3cnetSparseSemantic2021a, hanMultipathSensorySubstitution2025}.

Traditional LiDAR-based SSC methods \cite{chengS3cnetSparseSemantic2021a, liLODELocallyConditioned2023, leeSemcitySemanticScene2024} achieve high geometric accuracy due to high-quality 3D sensors. However, they suffer from high sensor costs, sparse points at long distances, and limited scalability for large-scale deployment. These limitations have driven research on camera-based SSC, which leverages dense image information to recover high-resolution 3D geometry and semantics without relying on expensive hardware \cite{caoMonosceneMonocular3d2022a}.

Most existing vision-based SSC methods follow a standard pipeline: first extract 2D features, then project to 3D voxels, and finally decode the voxels. In this paradigm, 2D features extracted by a backbone network are lifted into the 3D voxel space, and decoders predict occupancy and semantic labels. To reduce depth ambiguity, many methods introduce auxiliary depth estimation \cite{liVoxformerSparseVoxel2023a, jiangSymphonize3dSemantic2024} to provide coarse geometric priors within the camera frustum. Despite these advances, single-view SSC remains highly challenging due to severe occlusions, complex spatial dependencies, and uneven information density between visible and occluded regions.

It is noteworthy that many vision-based Semantic Scene Completion (SSC) methods rely on multi-frame sequences to extend the coverage of occluded areas, thereby reducing the difficulty of predicting hidden geometry. In contrast, SSC with single-view RGB input lacks temporal redundancy, which significantly increases the ill-posedness of accurately reconstructing occluded voxels.

Existing methods, such as MonoScene \cite{caoMonosceneMonocular3d2022a} and VoxFormer \cite{liVoxformerSparseVoxel2023a}, typically treat all voxels uniformly, ignoring the inherent physical differences between observed and unobserved regions. VisHall3D \cite{lu2025vishall3d} partially addresses this by partitioning visible and occluded regions at the structural level; however, it only encodes visibility as a geometric feature rather than enforcing it as a supervisory signal. As a result, the optimization dynamics remain entangled, and gradients from unreliable occluded regions can corrupt the reconstruction of visible surfaces.In contrast, VOIC introduces an explicit visibility-aware supervision mechanism via VRLE, which isolates visible-region supervision and guides global scene completion with more reliable visible priors.

To address this challenge, we propose VOIC, which introduces the Visible Region Label Extraction (VRLE) strategy to provide explicit and targeted supervision signals for the visible decoder. This approach keeps the learned visible-region features stable and reliable, achieving synergistic decoupling across both architectural and supervisory dimensions. VRLE generates visible voxel labels from complete 3D annotations, enabling a clear distinction between visible and occluded voxels and providing structured guidance for subsequent completion.

Building upon VRLE, VOIC employs a dual-decoder architecture: the Visible Decoder (VD) is supervised by VRLE-generated labels to precisely reconstruct the geometry and semantics of observed voxels; the Occlusion Decoder (OD) collaborates with the VD by utilizing normalized visible features as spatial-semantic priors. Furthermore, the OD provides global contextual feedback to optimize visible-region predictions while holistically inferring the entire scene to generate final complete voxel predictions.

To achieve high-precision geometric priors and effective feature interaction, VOIC integrates multi-level positional encoding and adaptive feature interaction modules. During the 3D projection stage, the Visible Embedding Feature Constructor (VEFC) injects spatial geometric information into 2D features. Through a dynamic learnable sampling mechanism, VEFC mitigates feature dilution associated with depth-estimation errors in traditional projection methods.

Extensive experiments on SemanticKITTI \cite{behleySemantickittidatasetsemantic2019a} and SSCBench-KITTI-360 \cite{liSscbenchLargescale3d2024} demonstrate that VOIC achieves strong and competitive performance across multiple evaluation metrics. The main contributions of this work are summarized as follows:

\begin{enumerate} 
\item  We propose VOIC, a novel single-view RGB-based SSC framework with external depth guidance that incorporates an offline Visible Region Label Extraction (VRLE) strategy to distinguish between visible and occluded voxels. This strategy enables a dual-decoder architecture, including a Visible Decoder (VD) and an Occlusion Decoder (OD), for collaborative reasoning: the VD establishes robust visible priors under explicit VRLE supervision, while the OD leverages these priors along with global supervision to infer the complete scene.
\item We introduce the Visible Embedding Feature Constructor (VEFC) and a multi-level positional encoding mechanism to systematically enhance voxel feature geometric discriminability and semantic alignment, providing a solid foundation for VD/OD collaborative reasoning.
\item Extensive experiments on SemanticKITTI and SSCBench-KITTI-360 validate the effectiveness of VOIC. Results show that the proposed method achieves competitive performance in both geometric completion and semantic segmentation.
\end{enumerate}


\section{Related Work}

\subsection{3D Input-based Approaches}

Early SSC methods leveraged native 3D data for high-fidelity reconstruction. The pioneering work, SSCNet~\cite{songSemanticscenecompletion2017b}, introduced SSC for indoor RGB-D scenes. Subsequent methods, such as 3DSketch~\cite{chen3dsketchawaresemantic2020a}, the approach by Li et al.~\cite{liRgbdbaseddimensional2019}, and the Cascaded Context Pyramid Network (CCPNet)~\cite{zhangCascadedcontextpyramid2019}, adopted the Truncated Signed Distance Function (TSDF) to better guide the prediction of geometry and semantics. 

The emergence of large-scale outdoor datasets, such as SemanticKITTI~\cite{behleySemantickittidatasetsemantic2019a} and OpenOccupancy~\cite{wangOpenoccupancylargescale2023}, subsequently catalyzed the development of methods tailored for outdoor environments.
A common strategy in this domain is to directly process LiDAR point clouds. For instance, LMSCNet~\cite{roldaoLmscnetLightweightmultiscale2020a} utilizes a 2D UNet to extract Bird's-Eye View (BEV) features from sparse scans, which are then unrolled along the height dimension to complete the 3D scene. JS3C-Net \cite{yanSparsesinglesweep2021a} proposes a sparse single-sweep segmentation framework with contextual shape priors and leverages multi-frame LiDAR aggregation to convert sparse scans into dense representations.

To leverage complementary information from images and LiDAR, recent methods focus on multi-modal fusion. Chang et al. \cite{changMultiphasecameraLiDARfusion2023} address unreliable image features and modality trade-offs by introducing a weakly-supervised loss, attention-based feature fusion, and confidence-aware late fusion. IPVoxelNet \cite{luLiDARcameracontinuousfusion2024} maps images and point clouds into a unified voxel space, independently learns geometric and semantic features, and uses cross-modal knowledge distillation to enhance scene understanding.

While 3D input-based methods achieve remarkable accuracy, their dependency on expensive sensors like LiDAR limits practical scalability. This constraint has motivated a parallel line of research into vision-centric methods, which seek to infer complete 3D scenes from low-cost, readily available monocular or multi-view images.

\subsection{Visual-Centric Approaches}

Vision-centric SSC methods address the challenge of reconstructing 3D scenes from images. The single-image SSC pipeline was largely established by MonoScene~\cite{caoMonosceneMonocular3d2022a}, which lifts 2D image features into a 3D frustum to generate dense voxel representations. Early innovations primarily focused on enhancing the 3D decoder: OccFormer~\cite{zhangOccformerDualpathTransformer2023a} provides efficient, long-range encoding for voxel features; NDC-scene~\cite{yaoNdcsceneBoostMonocular2023a} projects 2D features into Normalized Device Coordinate (NDC) space to recover depth; and TPVFormer~\cite{huangTriperspectiveViewVisionbased2023a} represents 3D space with tri-perspective views and uses Transformers to capture spatial dependencies. HASSC~\cite{wangNotallvoxels2024a} tackles the class imbalance issue via a hard sample-aware training strategy that dynamically selects challenging voxels for refinement.

A core challenge in single-image SSC is depth ambiguity. VoxFormer~\cite{liVoxformerSparseVoxel2023a} systematically addresses this by first generating sparse depth estimates and then using deformable attention to propagate and densify contextual features. Beyond geometry, instance-level reasoning has also been integrated to boost performance. Symphonies~\cite{jiangSymphonize3dSemantic2024} employs serial instance propagation attention to model instance semantics and reduce occlusion errors, while Xiao et al.~\cite{xiaoInstanceawaremonocular3D2024} propose an instance-aware framework that fuses mask-derived cues via a dedicated attention module.

The field has recently seen significant advances through multi-modal fusion and temporal modeling. MiXSSC~\cite{wangMixsscForwardbackwardmixture2025} combines sparse features from forward projection with dense depth priors from backward projection. HTCL~\cite{liHierarchicalTemporalContext2025a} introduces hierarchical temporal context learning for better consistency, CurriFlow~\cite{linCurriFlowCurriculumGuidedDepth2025} leverages optical flow and curriculum learning, and CF-SSC~\cite{luOneStepCloser2025} extends the perceptual range via pseudo-future frame prediction.

Architectural innovations continue to push the boundaries. FoundationSSC~\cite{chenUnleashingSemanticGeometric2025} uses stereo (binocular) input and proposes a dual-decoupling framework for separate semantic and geometric refinement. MVFormer~\cite{gaoMVFormerUNetlikeTransformer2025} utilizes a UNet-like decoder with Mix-Voxel attention, VFG-SSC~\cite{phamSemisupervised3DSemantic2025} adopts a semi-supervised approach to leverage 2D foundation models, SPHERE~\cite{yangSPHERESemanticPHysicalEngaged2025} jointly models semantics and physics with voxels and Gaussians, and MARE~\cite{tsengMemoryAugmentedReCompletion3D2025} introduces a memory bank to complete unseen regions. VisHall3D~\cite{lu2025vishall3d} introduces a “Vision vs. Hallucination” strategy to separate visible-region reconstruction from occluded-region completion,  but it does not explicitly construct label-level visibility-aware supervision targets in the manner of VRLE.

In summary, while many vision-centric methods mitigate occlusion by leveraging multi-frame temporal redundancy, such approaches do not address the fundamental issue of supervision ambiguity in single-image settings. To bridge this gap, we propose VOIC, which explicitly reformulates SSC through a supervision-decoupling strategy. By introducing distinct learning objectives for visible-region perception and occluded-region prediction, our method mitigates supervision contamination and enables structured, high-fidelity scene completion for single-view RGB-based SSC with external depth guidance and without temporal aggregation.

\begin{figure}[t]
    \centering
    \includegraphics[width=\columnwidth]{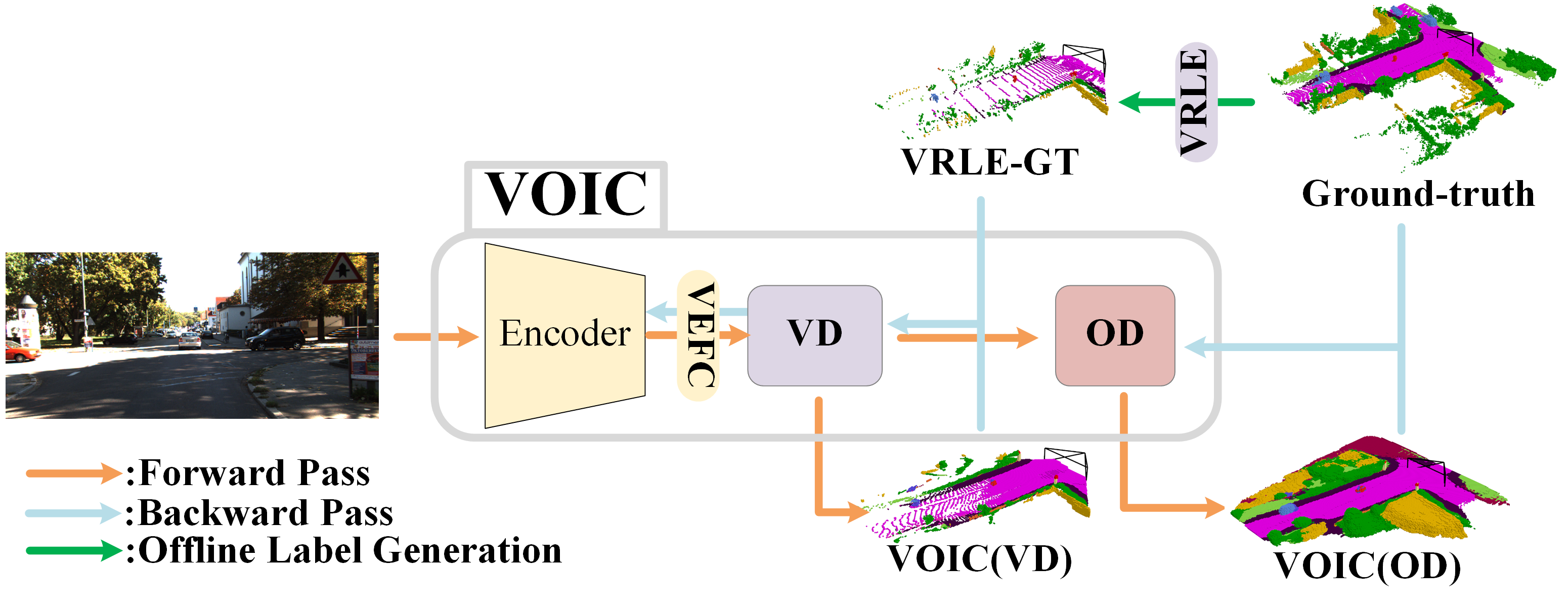}
    \caption{Overall framework of the proposed method. During training, the reference RGB image and the external depth prior are processed by the backbone and VEFC to construct the visible embedding. VD predicts visible voxels under VRLE supervision, while OD predicts the complete scene under GT supervision. During inference, only the forward pass is performed.}
    \label{fig:VOIC}
  
\end{figure}

\begin{figure*}[htb!]
    \centering
    \includegraphics[width=\textwidth]{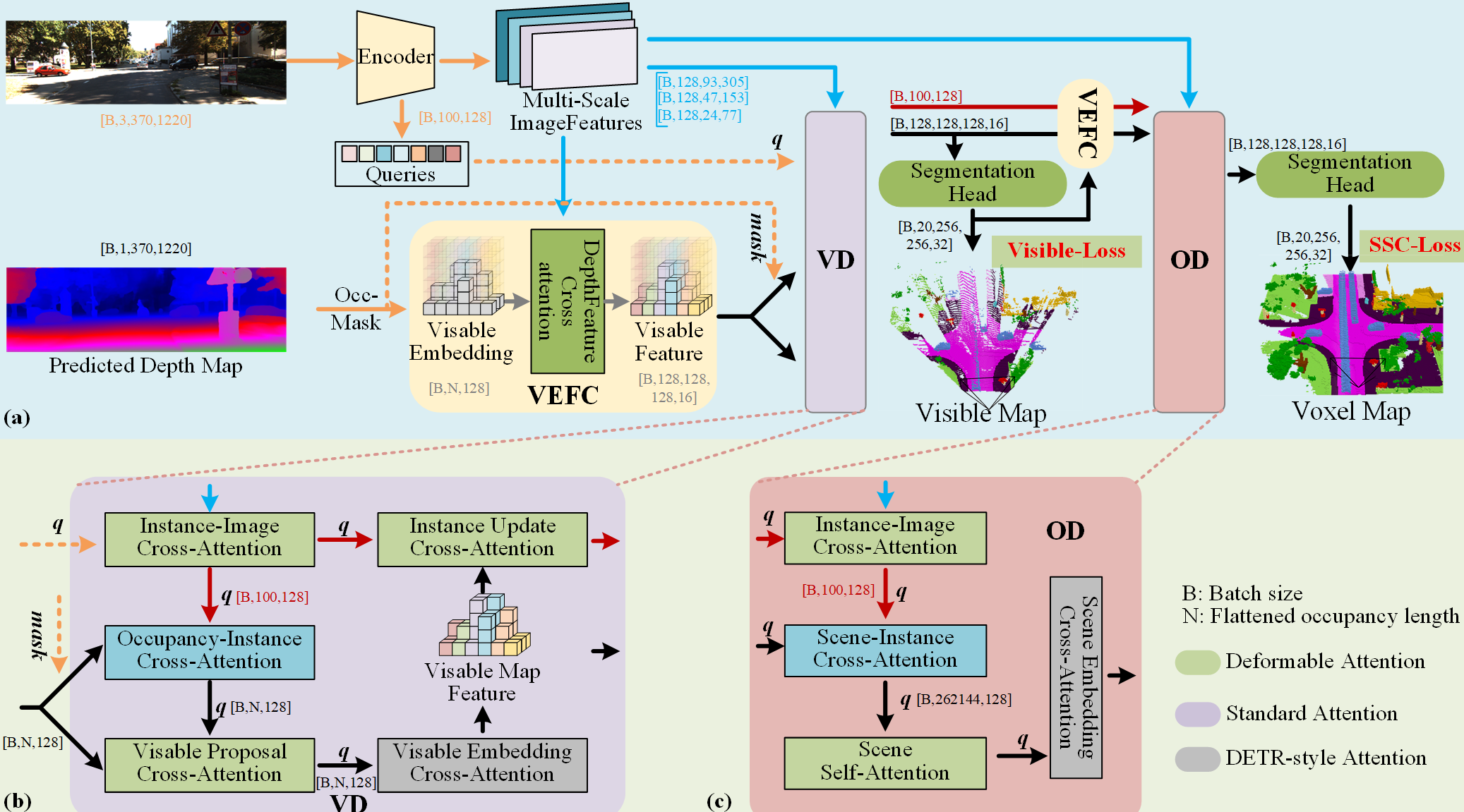}
\caption{
Overall architecture of the VOIC framework. 
(a) The model follows a progressive visible–occluded paradigm that decouples the $3$D SSC task through differentiated supervision.
The Visible Embedding Feature Constructor (VEFC) lifts $2$D image features into $3$D and fuses them with depth-derived occupancy to form a unified volumetric representation. 
(b) The Visible Decoder (VD) predicts geometry and semantics of observed regions under explicit visible-voxel supervision. 
(c) The Occlusion Decoder (OD) leverages normalized visible features as spatial–semantic priors and reconstructs the full $3$D scene under global scene completion supervision.
}
    \label{fig:fig-model}

\end{figure*}

\section{Methodology}

\subsection{Overview}
We introduce VOIC, a unified framework designed to address a critical yet underexplored issue in 3D Semantic Scene Completion (SSC): inaccurate supervision on visible regions can systematically degrade global scene completion. Formally, given a reference RGB image $I^{\text{rgb}} \in \mathbb{R}^{H_{\text{img}} \times W_{\text{img}} \times 3}$ and an externally predicted depth prior $\mathbf{D}_{\text{pre}}$, the objective is to predict the complete scene geometry and semantics $\hat{\mathbf{Y}}$, which approximates the ground truth $\mathbf{Y} \in \mathbb{R}^{X \times Y \times Z \times N_c}$. This is formulated as learning a network $\Theta$ such that $\hat{\mathbf{Y}} = \Theta(I^{\text{rgb}}, \mathbf{D}_{\text{pre}})$.

To facilitate this decoupled learning paradigm, our framework incorporates a dual supervision mechanism:
\begin{enumerate}
    \item Explicit Visibility Supervision: Derived via an offline Visible Region Label Extraction (VRLE) strategy (Sec.~\ref{sec:vrle}), this supervision guides the network to capture precise features of observable surfaces.
    \item Holistic Scene Supervision: Utilizing the complete ground-truth annotations, this forces the network to infer occluded regions based on the established visible priors.
\end{enumerate}

The overall computational pipeline is illustrated in Figure~\ref{fig:VOIC}. VRLE-generated labels provide supervision for the Visible Decoder (VD), while global ground-truth labels supervise the Occlusion Decoder (OD).
During the forward pass, as illustrated in Figure \ref{fig:fig-model}, the pipeline first employs the Visible Embedding Feature Constructor (VEFC) (Sec.~\ref{sec:vefc}) to lift 2D image features into a 3D volumetric representation. Specifically, VEFC leverages a ResNet-50 backbone \cite{liMaskdinounified2023} and a DETR-style attention module \cite{zhuDeformableDETRDeformable2021} to strengthen the coupling between image cues and 3D occupancy.

These 3D features are subsequently processed by the Visible Decoder (VD) (Sec.~\ref{sec:vd}) to reconstruct the geometry and semantics of observable surfaces under visibility supervision. Building upon VD’s high-fidelity predictions, the Occlusion Decoder (OD) (Sec.~\ref{sec:od}) uses these results as spatial–semantic priors to infer the complete 3D scene under holistic supervision, ensuring a coherent reconstruction. Finally, a segmentation head with an ASPP module \cite{chenDeeplabSemanticimage2017} produces the final per-voxel semantic predictions.

\subsection{Visible Region Label Extraction (VRLE)}
\label{sec:vrle}

\begin{figure}[t]
    \centering
    \includegraphics[width=\columnwidth]{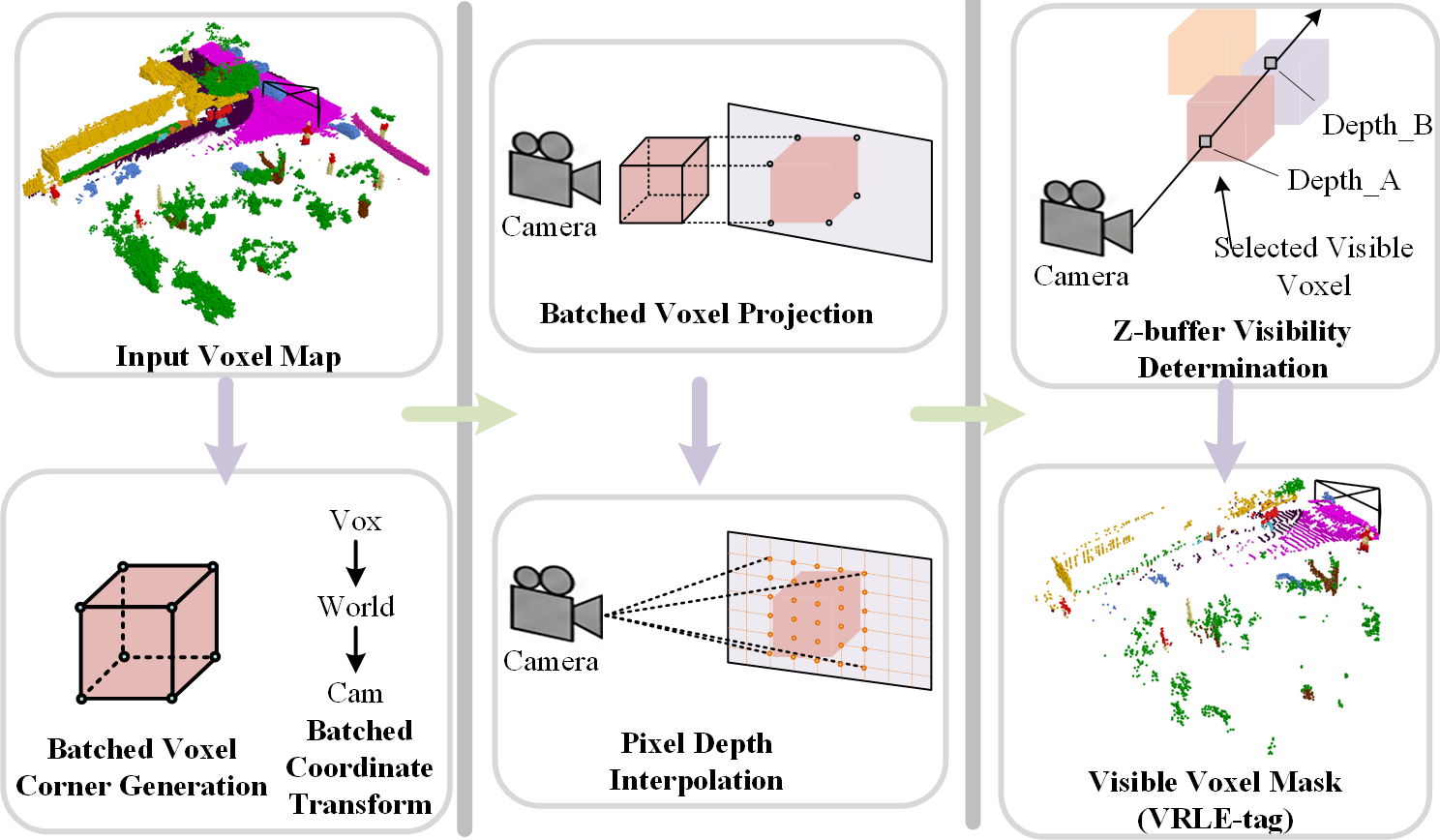}
    \caption{Overview of VRLE module.}
    \label{fig:VRLE}
\end{figure}

Standard SSC datasets (e.g., SemanticKITTI \cite{behleySemantickittidatasetsemantic2019a}) provide complete 3D semantic annotations but inherently lack explicit distinctions between visible and occluded voxels. This uniform supervision fails to account for the fundamentally different nature of visible perception versus occluded reasoning. To solve this, we introduce VRLE, an offline label generation strategy that provides unambiguous supervision for the proposed Visible Decoder by performing a visibility-aware projection. The computational process is illustrated in Figure~\ref{fig:VRLE}.
The VRLE process begins with \textit{Voxel Instantiation and Geometric Projection}. Given the complete ground-truth semantic voxel grid 
$\mathbf{Y} \in \mathbb{R}^{X \times Y \times Z \times N_c}$, 
we first extract the set of occupied voxels
\begin{equation}
\mathcal{V} = \left\{ v_i \mid Y_i \neq \text{empty} \right\},
\end{equation}
where $N_c$ denotes the number of semantic classes, including the empty class.
 For each voxel $v_i$, we define its structure by eight vertices $\mathcal{P}_i^{\text{vox}} \in \mathbb{R}^{8 \times 3}$. Since visibility is determined by the voxel’s actual surfaces rather than its center, representing each voxel with eight vertices allows explicit modeling of surface-level occlusion. 
To simulate the imaging process and establish correspondence between the 3D voxel structure and the 2D input image, each voxel vertex $\mathbf{p}_k^{\text{vox}}$ is transformed to the camera coordinate system and then projected onto the image plane using the standard pinhole camera model. This pipeline uses the voxel-to-world transformation $\mathbf{T}_{v \to w}$, the camera extrinsic parameters $[\mathbf{R}|\mathbf{t}]$, and the intrinsic matrix $\mathbf{K}$, and is formally defined as:

\begin{equation}
\label{eq:projection}
\begin{gathered}
\mathbf{p}_k^{\text{world}} =
\mathbf{T}_{v \to w}\mathbf{p}_k^{\text{vox}}\\
\mathbf{p}_k^{\text{cam}} =
\mathbf{R}\mathbf{p}_k^{\text{world}} + \mathbf{t} \\
u =
\operatorname{round}\!\left(
f_x \frac{x}{z} + c_x
\right), \quad
v =
\operatorname{round}\!\left(
f_y \frac{y}{z} + c_y
\right),
\end{gathered}
\end{equation}

where $\mathbf{p}_k^{\text{cam}}=[x,y,z]^\top$ denotes the point in the camera coordinate frame, $f_x,f_y$ are the focal lengths, and $(c_x,c_y)$ is the principal point. The projected vertices form a 2D polygon $\Pi_i$ corresponding to voxel $v_i$.

Identifying truly visible voxels requires eliminating depth ambiguity due to occlusions. Since traditional ray-casting is prohibitively slow for dense voxel grids, we employ a \textit{Vectorized Sparse Rasterization pipeline enhanced with Z-buffering}. For computational efficiency, we first perform Sparse Bounding-Box Sampling: for each projected voxel $\Pi_i$, we compute its 2D Axis-Aligned Bounding Box (AABB) $\mathcal{B}_i$ and generate a sparse grid of candidate pixels $\mathcal{S}_i$ within $\mathcal{B}_i$ using a sampling stride $\delta$ (e.g., $\delta=4$).
Next, in the Inclusion Verification and Depth Interpolation stage, we decompose the cube's projected volume into its six quadrilateral faces. A vectorized cross-product test is employed to determine if a candidate pixel $p \in \mathcal{S}_i$ falls within any projected face. For every validated pixel $p$, we calculate its precise depth $d_p$ through robust linear interpolation of the face vertices' depths, explicitly handling degenerate projection cases to ensure numerical stability.
We maintain a global depth buffer $\mathcal{Z} \in \mathbb{R}^{\text{H}_{img} \times \text{W}_{img}}$ to record the minimum depth. The visibility mask is generated by the following rule: a voxel $v_i$ is marked as visible if its depth is the smallest at any covered pixel location $(u, v)$:
\vspace{-4pt}
\begin{equation}
\label{eq:visibility}
v_i \in \mathcal{V}_{\text{vis}} \iff \exists (u,v) \in \Pi_i, \text{ s.t. } d_{u,v}^{(i)} = \min_{j \in \mathcal{V}} (d_{u,v}^{(j)}),
\vspace{-6pt}
\end{equation}
where $d_{u,v}^{(j)}$ is the interpolated depth of voxel $v_j$ at pixel $(u, v)$.
Finally, VRLE generates a binary visibility mask $\mathbf{M}_{\text{vis}}$. To suppress moving-object artifacts inherent in the SemanticKITTI temporal aggregation,
we project the predicted depth map $D_{\text{pre}}$ into 3D space to obtain an auxiliary
occupancy mask $\mathbf{M}$. We then derive the visible targets as
\begin{equation}
\mathbf{Y}_{\text{vis}} = \mathbf{Y} \cap \left( \mathbf{M}_{\text{vis}} \cup \mathbf{M} \right).
\end{equation} This $\mathbf{Y}_{\text{vis}}$ provides dedicated supervision for the Visible Decoder, while the complete ground truth $\mathbf{Y}$ supervises the Occlusion Decoder to ensure holistic scene understanding.

\subsection{Visible Embedding Feature Constructor (VEFC)}
\label{sec:vefc}

VEFC constructs a geometrically-grounded scene embedding by explicitly mapping image appearances onto 3D structures, thereby suppressing unreliable hallucinations. Specifically, we project the predicted depth map $\mathbf{D}_{\text{pre}}$ into world coordinates to generate a binary occupancy mask $\mathbf{M} \in \{0, 1\}^{X \times Y \times Z}$, identifying physically present voxels. Unlike methods relying on unconstrained learnable queries, we initialize content-free queries $\mathbf{Q}_{\text{content}}$ only at valid voxel locations indicated by $\mathbf{M}$ and fuse them with image features $\mathbf{F}_{\text{2D}}$ via Deformable Attention:
\begin{equation}
\label{eq:deform_attn}
\mathbf{F}_{\text{voxel}} = \text{DeformAttn}(\mathbf{Q}_{\text{content}} + \mathbf{S}_{\text{pos}}, \mathcal{P}(\mathbf{S}_{\text{pos}}), \mathbf{F}_{\text{2D}}),
\end{equation}
where $\mathcal{P}$ denotes the 3D-to-2D projection and $\mathbf{S}_{\text{pos}}$ represents the 3D geometric positional embedding. By enforcing this explicit 3D-image correspondence, VEFC ensures that the resulting voxel features are conditioned on depth-derived visible geometry. This mechanism effectively prevents the "hallucinated" responses often generated by free-form queries in occluded or complex regions, providing a deterministic foundation for subsequent scene completion.

\begin{figure}[htbp]
  \centering
  \includegraphics[width=\columnwidth]{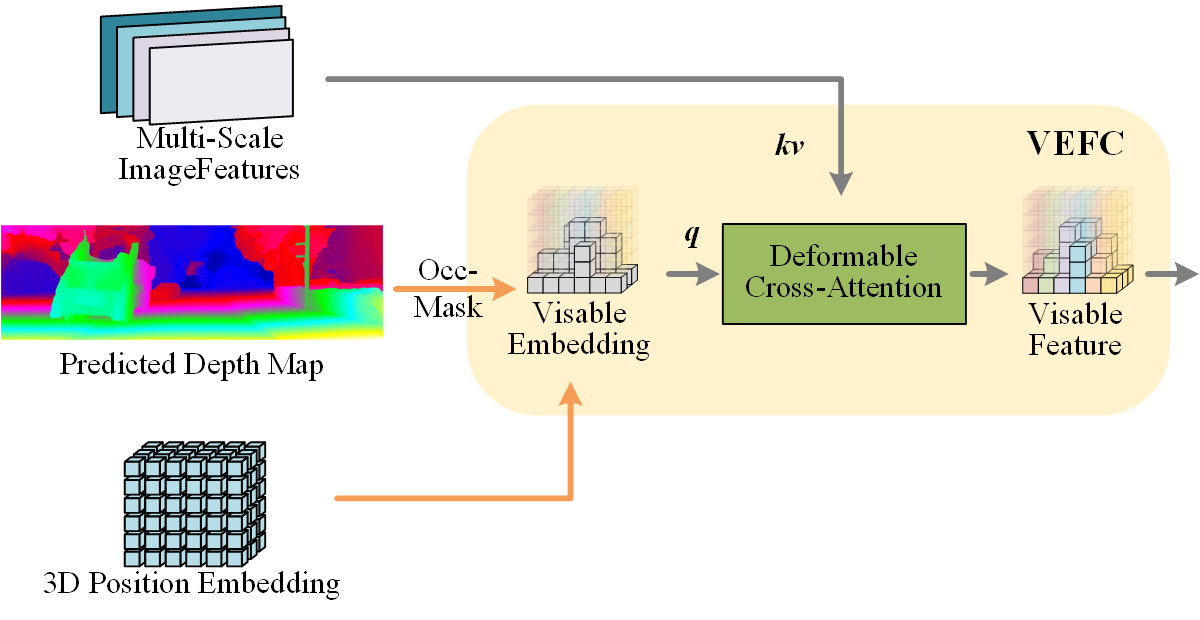}
  \caption{Sparse Voxel Feature Initialization. 
  The VEFC module creates a geometry-driven 3D representation using a zero-initialized query. An occupancy mask sparsifies the structure, while 3D positional encoding enables precise localization. Deformable Attention then aggregates 2D image features into the visible voxels.}
  \label{fig:vefc}
\end{figure}

\subsection{Visible Decoder (VD)}
\label{sec:vd}

As illustrated in Fig.~\ref{fig:fig-model}(b), the Visible Decoder (VD) serves as the primary component for achieving high-fidelity semantic perception within the visible regions. The comprehensive data flow and feature transformation procedures are formally summarized in Algorithm~\ref{alg:vd_forward}.

The VD architecture is designed to balance computational efficiency with structural robustness through a multi-stage integration paradigm. First, to mitigate the memory overhead associated with high-resolution multi-scale image features $\mathbf{F}_{\text{2D}}$, we employ a set of instance queries $\mathbf{Q}_{\text{inst}}$ to compress and aggregate the most salient visual cues. These compressed queries subsequently interact with the critical voxels identified by the occupancy mask $\mathbf{M}$ through the \textit{Occupancy-Instance Cross Attention} mechanism. This step facilitates the explicit fusion of image-derived semantics with the skeleton of the 3D geometric structure.

Considering that the initial occupancy mask $\mathbf{M}$—derived from depth estimation—may contain inherent inaccuracies or misalignments, we incorporate a structural refinement strategy. The sparse features obtained from the masked interaction are re-integrated with the holistic (unmasked) voxel features $\text{Flatten}(\mathbf{F}_{\text{vis}})$, where $\mathbf{F}_{\text{vis}}$ is initialized from the VEFC output $\mathbf{F}_{\text{voxel}}$ and represents the full visible-region voxel representation.  By leveraging the global 3D scene context, this process effectively compensates for potential errors in the mask, ensuring spatial consistency across the observable volume. To further enrich the categorical representation, the resulting features are fused with class embedding priors $\mathbf{C}_{\text{vd}}$ via a DETR-style attention module, yielding the final visible-region output features $\mathbf{F}_{\text{vis}}^{\text{VD}}$.

A key characteristic of our design is the reciprocal guidance between the scene features and the queries. Since $\mathbf{F}_{\text{vis}}^{\text{VD}}$ encapsulates refined occupancy information and semantic priors, it is utilized to update the instance queries $\mathbf{Q}_{\text{inst}}'$  through 3D Deformable Attention. This feedback loop ensures that the updated queries $\mathbf{Q}_{\text{inst}}^{\text{VD}}$  carry explicit 3D-aware constraints. Consequently, in the subsequent Occlusion Decoder (OD), these queries can establish more precise correspondences with image features, providing superior guidance for reasoning within the unobserved spaces.

\begin{algorithm}[t]
\caption{Forward Pass of the Visible Decoder (VD)}
\label{alg:vd_forward}
\begin{algorithmic}[1]
\Require 
    Initial 3D embedding for VD from VEFC $\mathbf{F}_{\text{vis}}$ (Shape: $B \times C \times X \times Y \times Z$), 
    2D multi-scale image features $\mathbf{F}_{\text{2D}}$, 
    Instance queries $\mathbf{Q}_{\text{inst}}$ (Shape: $B \times N_q \times C$), 
    Occupancy mask $\mathbf{M}$, 
    2D instance reference points $\mathbf{R}_{\mathrm{2D}}$,
    3D voxel reference points $\mathbf{R}_{\mathrm{3D}}$,
    Class embeddings $\mathbf{C}_{\text{vd}}$.
\Ensure 
    Refined visible features $\mathbf{F}_{\text{vis}}^{\text{VD}}$, 
    Updated instance queries $\mathbf{Q}_{\text{inst}}''$.

\State Flatten $\mathbf{F}_{\text{2D}}$ into 1D sequences;
\State $\mathbf{Q}_{\text{inst}}' \leftarrow \text{DeformAttn}(\mathbf{Q}_{\text{inst}}, \mathbf{R}_{\text{2D}}, \mathbf{F}_{\text{2D}})$; \Comment{Instance-Image Cross-Attn, output shape: $B \times N_q \times C$}

\State $\mathbf{E}_{\text{vis}}^{\text{flat}} \leftarrow \text{Flatten}(\mathbf{F}_{\text{vis}}) [\mathbf{M}]$; \Comment{Output: $B \times N \times C$, $N$ is valid voxels}

\State $\mathbf{E}_{\text{vis}}^{\text{flat}} \leftarrow \text{CrossAttn}(\mathbf{E}_{\text{vis}}^{\text{flat}}, \mathbf{Q}_{\text{inst}}')$; \Comment{Occupancy-Instance Cross-Attn}
\State $\mathbf{E}_{\text{vis}}^{\text{flat}} \leftarrow \text{DeformAttn}(\mathbf{E}_{\text{vis}}^{\text{flat}}, \mathbf{R}_{\text{3D}}, \text{Flatten}(\mathbf{F}_{\text{vis}}))$; \Comment{Visible Proposal Cross-Attn}
\State $\mathbf{E}_{\text{vis}}^{\text{final}} \leftarrow \text{CrossAttn}(\mathbf{E}_{\text{vis}}^{\text{flat}}, \mathbf{C}_{\text{vd}})$; \Comment{Visible Embedding Cross-Attn}

\State $\mathbf{F}_{\text{vis}}^{\text{VD}} \leftarrow \text{IndexBack}(\mathbf{0}, \mathbf{E}_{\text{vis}}^{\text{final}}, \mathbf{M})$; \Comment{Scatter sparse features onto a zero-initialized grid}
\State $\mathbf{Q}_{\text{inst}}^{\text{VD}} \leftarrow \text{DeformAttn}(\mathbf{Q}_{\text{inst}}', \mathbf{R}_{\text{3D}}, \text{Flatten}(\mathbf{F}_{\text{vis}}^{\text{VD}}))$; \Comment{Instance Update Cross-Attn}

\State \Return $\mathbf{F}_{\text{vis}}^{\text{VD}}$, $\mathbf{Q}_{\text{inst}}^{\text{VD}}$
\end{algorithmic}

\end{algorithm}

\subsection{Occlusion Decoder (OD)}
\label{sec:od}

The Occlusion Decoder (OD) is designed to reconstruct the complete 3D scene by propagating structural and semantic priors from the visible regions into the unobserved spaces. The core intuition of the OD lies in leveraging the high-confidence visible-region features to guide the reasoning of hidden structures through a multi-stage feature evolution paradigm. 

First, to enhance the semantic understanding of the queries, the instance queries $\mathbf{Q}_{\text{inst}}^{\text{VD}}$ derived from the VD—which already encapsulate preliminary 3D spatial awareness—are re-fused with the multi-scale image features $\mathbf{F}_{\text{2D}}$ through \textit{Instance-Image Cross-Attention}. By re-grounding the queries in the image plane, the model strengthens its grasp of the object-level semantics before initiating 3D completion. Simultaneously, the refined visible-region features $\mathbf{F}_{\text{vis}}^{\text{VD}}$ are injected back into the holistic grid via the IndexBack operation using the occupancy mask $\mathbf{M}$. This ensures that the established visible priors serve as fixed anchors within the integrated representation $\mathbf{E}_{\text{occ}}$, providing reliable spatial guidance for the subsequent completion process.

Subsequently, the OD achieves cross-modal integration by fusing the image-enhanced queries $\mathbf{Q}_{\text{inst}}^{\text{OD}}$ with the initial 3D grid $\mathbf{E}_{\text{occ}}$ through \textit{Scene-Instance Cross-Attention}. To further solidify the feature representation, a \textit{Scene Self-Attention} mechanism is employed. By performing deformable spatial reasoning over the entire volume, the model enhances the internal expressive power of the features, allowing structural cues to propagate seamlessly from observable boundaries to occluded regions. 

Finally, the categorical expressive capability is augmented by the \textit{Scene Embedding Cross-Attention} stage. Here, a specialized set of class-level embeddings $\mathbf{C}_{\text{od}}$ interacts with the refined voxel features via a DETR-style decoupled attention module. This process fuses global scene priors with high-dimensional category information, ensuring that the completed scene is both geometrically consistent and semantically plausible. The OD ultimately outputs the fully refined voxel features $\mathbf{F}_{\text{occ}}^{\text{OD}}$, which are then passed to the segmentation head for the final holistic scene prediction.

\begin{algorithm}[t]
\caption{Feature Propagation in Occlusion Decoder (OD)}
\label{alg:od_forward}
\begin{algorithmic}[1]
\Require 
    Initial 3D embedding for OD from VEFC $\mathbf{F}_{\text{occ}}$, 
    Visible-refined features $\mathbf{F}_{\text{vis}}^{\text{VD}}$, 
    Instance queries $\mathbf{Q}_{\text{inst}}^{\text{VD}}$, 
    2D image features $\mathbf{F}_{\text{2D}}$, 
    Occupancy mask $\mathbf{M}$, 
    2D instance reference points $\mathbf{R}_{\mathrm{2D}}$,
    3D voxel positions $\mathbf{R}_{\mathrm{3D}}$,
    Class embeddings $\mathbf{C}_{\text{od}}$.
\Ensure 
    Holistic refined features $\mathbf{F}_{\text{occ}}^{\text{OD}}$.

\State $\mathbf{Q}_{\text{inst}}^{\text{OD}} \leftarrow \text{DeformAttn}(\mathbf{Q}_{\text{inst}}^{\text{VD}}, \mathbf{R}_{\text{2D}}, \mathbf{F}_{\text{2D}})$; \Comment{Instance-Image Cross-Attn}

\State $\mathbf{E}_{\text{occ}} \leftarrow \text{IndexBack}(\mathbf{F}_{\text{occ}}, \mathbf{F}_{\text{vis}}^{\text{VD}}, \mathbf{M})$; \Comment{Inject VD priors into the OD grid}
\State $\mathbf{E}_{\text{occ}}^{\text{flat}}, \text{shape} \leftarrow \text{Flatten}(\mathbf{E}_{\text{occ}})$;

\State $\mathbf{E}_{\text{occ}}^{\text{flat}} \leftarrow \text{CrossAttn}(\mathbf{E}_{\text{occ}}^{\text{flat}}, \mathbf{Q}_{\text{inst}}^{\text{OD}})$; \Comment{Scene-Instance Cross-Attn}
\State $\mathbf{E}_{\text{occ}}^{\text{flat}} \leftarrow \text{DeformAttn}(\mathbf{E}_{\text{occ}}^{\text{flat}}, \mathbf{R}_{\text{3D}}, \mathbf{E}_{\text{occ}}^{\text{flat}})$; \Comment{Geometric completion via Self-Attn}

\State $\mathbf{E}_{\text{occ}}^{\text{final}} \leftarrow \text{CrossAttn}(\mathbf{E}_{\text{occ}}^{\text{flat}}, \mathbf{C}_{\text{od}})$; \Comment{Scene Embedding Cross-Attn}
\State $\mathbf{F}_{\text{occ}}^{\text{OD}} \leftarrow \text{Reshape}(\mathbf{E}_{\text{occ}}^{\text{final}}, \text{shape})$; \Comment{Restore to 3D volume}

\State \Return $\mathbf{F}_{\text{occ}}^{\text{OD}}$
\end{algorithmic}

\end{algorithm}

\subsection{Loss Functions}
\label{sec:loss}

The overall training objective is defined as:
\begin{equation}
\label{eq:total_loss}
L_{\text{total}} = \lambda_{\text{VD}} L_{\text{VD}} + \lambda_{\text{OD}} L_{\text{OD}},
\end{equation}
where $\lambda_{\text{VD}}$ and $\lambda_{\text{OD}}$ are trade-off hyperparameters.To balance visible-region reconstruction and occluded-region completion in holistic scene understanding, we empirically set $\lambda_{\text{VD}} = \lambda_{\text{OD}} = 1.0$.

Both decoders are optimized using the same objective, which jointly supervises geometric occupancy and semantic prediction:
\begin{equation}
\label{eq:loss_X}
L_{X} = \lambda_{\text{scal}} L_{\text{scal}}^{\text{geo}} + \lambda_{\text{ce}} L_{\text{ce}} + \lambda_{\text{mIoU}} L_{\text{mIoU}},
\end{equation}
where $X \in \{\text{VD}, \text{OD}\}$. $L_{\text{scal}}^{\text{geo}}$ follows MonoScene~\cite{caoMonosceneMonocular3d2022a} and $L_{\text{ce}}$ is the voxel-wise cross-entropy loss used in Symphonies~\cite{jiangSymphonize3dSemantic2024}. $\lambda_{\text{scal}}$, $\lambda_{\text{ce}}$, and $\lambda_{\text{mIoU}}$ denote weighting coefficients.

To better align optimization with the evaluation metric, a differentiable semantic mIoU loss is adopted. Let $\Omega$ denote the set of valid voxels (excluding ignore labels) within the holistic grid $\mathbb{R}^{X \times Y \times Z}$. For each voxel $i \in \Omega$, let $\hat{y}_{i,c}$ and $y_{i,c}$ denote the predicted probability and the one-hot ground-truth label for class $c$, respectively. The soft IoU for class $c$ is defined as:
\begin{equation}
\label{eq:soft_iou}
\text{IoU}_c = \frac{\sum_{i \in \Omega} \hat{y}_{i,c} y_{i,c}}{\sum_{i \in \Omega} \hat{y}_{i,c} + \sum_{i \in \Omega} y_{i,c} - \sum_{i \in \Omega} \hat{y}_{i,c} y_{i,c}},
\end{equation}
where $c \in \{1, \dots, N_c-1\}$ (excluding the empty category).

The final mIoU loss is formulated as:
\begin{equation}
\label{eq:miou_loss}
L_{\text{mIoU}} = 1 - \frac{1}{N_c - 1} \sum_{c=1}^{N_c - 1} \text{IoU}_c,
\end{equation}


\section{Experiments}
\label{sec:experiments}

In this section, we present the experimental evaluations of our proposed method on the SemanticKITTI~\cite{behleySemantickittidatasetsemantic2019a} and SSCBench-KITTI-360~\cite{liSscbenchLargescale3d2024} datasets. 
Comparative results against existing state-of-the-art approaches are provided in Sec.~\ref{sec:comparison}, 
while comprehensive ablation studies are reported in Sec.~\ref{sec:ablation} to further analyze the effectiveness of each component.

\subsection{Dataset and Metric}
\label{sec:dataset}

We conduct experiments on the SemanticKITTI~\cite{behleySemantickittidatasetsemantic2019a} and SSCBench-KITTI-360~\cite{liSscbenchLargescale3d2024} datasets, both of which offer densely annotated urban driving scenes derived from the KITTI Odometry Benchmark. 
The scenes are voxelized into grids of size $256 \times 256 \times 32$, covering a spatial volume of $51.2\text{m} \times 51.2\text{m} \times 6.4\text{m}$ with a voxel size of 0.2\,m.

SemanticKITTI contains 10 training sequences, 1 validation sequence, and 11 hidden test sequences. It provides RGB images with a resolution of $1226 \times 370$ and includes 20 semantic classes.  
SSCBench-KITTI-360 provides 7 training sequences, 1 validation sequence, and 1 test sequence, offering RGB images of size $1408 \times 376$ and defining 19 semantic classes.

We follow standard SSC evaluation protocols and report Intersection over Union (IoU) and mean IoU (mIoU). 
IoU assesses binary occupancy (empty vs. occupied) and reflects geometric reconstruction performance, whereas mIoU evaluates per-class semantic accuracy—the primary metric in most SSC benchmarks\cite{caoMonosceneMonocular3d2022a}\cite{jiangSymphonize3dSemantic2024}\cite{liVoxformerSparseVoxel2023a}\cite{yangSPHERESemanticPHysicalEngaged2025}.

\subsection{Implementation Details}
\label{sec:implementation}

We train our model on two RTX A6000 GPUs with a batch size of 1.  
Random horizontal flipping\cite{zhangBeverseUnifiedperception2022}\cite{liBevdepthAcquisitionreliable2023}\cite{huangBEVDetHighperformanceMulticamera2022} and color jittering (brightness in the range $[1.2, 1.25]$, contrast in the range $[0.6, 0.65]$, and saturation adjustment of $0.1$) are applied as image augmentation methods.  

Following previous methods~\cite{liVoxformerSparseVoxel2023a,jiangSymphonize3dSemantic2024,meiCamerabased3dsemantic2024a,gaoMVFormerUNetlikeTransformer2025,yuContextgeometryaware2024,lu2025vishall3d}, we use an off-the-shelf MobileStereoNet~\cite{shamsafarMobilestereonetlightweightdeep2022} to generate depth priors. During both training and inference, the SSC network takes a reference RGB image together with the predicted depth prior as input.

Training is performed using the AdamW\cite{loshchilovDecoupledWeightDecay2019} optimizer with an initial learning rate of $3.5\times10^{-4}$, a weight decay of $0.015$, and momentum coefficients $(\beta_1, \beta_2) = (0.9, 0.99)$.  
The loss weights for both VD and OD are set as \(\lambda_{\text{scal}} = 1\), \(\lambda_{\text{ce}} = 1\), and \(\lambda_{\text{miou}} = 10\).

Following other comparable methods\cite{tsengMemoryAugmentedReCompletion3D2025}\cite{jiangSymphonize3dSemantic2024}, the ResNet-50\cite{heDeepresiduallearning2016} backbone and image encoder are initialized with pre-trained MaskDINO\cite{liMaskdinounified2023} weights.  

A MultiStepLR scheduler is adopted, where the learning rate is decayed by a factor of 0.1 at epochs 12 and 15. The total training length is 16 epochs.  

We set the input spatial voxel size to $128 \times 128 \times 16$ with $C=128$ channels. The voxel representation is upsampled to $256 \times 256 \times 32$ using a 3D transposed convolution.

\subsection{Main Results}
\label{sec:comparison}

\begin{table*}[ht]
\caption{Quantitative results on the SemanticKITTI hidden test set. $^\mathbf{T}$ represents the use of multi-frame input. $^\dagger$ represents the use of multi-camera images as input. The best and the second best results are marked in \textbf{bold} and $\underline{\text{underlined}}$, respectively.}
\label{tab:semantickitti_results}
\centering
\small
\resizebox{\linewidth}{!}{%
\begin{tabular}{l|cc|*{19}{c}}
\toprule
\textbf{Method} & \textbf{IoU} &\textbf{mIoU}&
\rotatebox{90}{\shortstack[l]{\textbf{\textcolor{road}{\rule{8pt}{8pt}} road}\\{\tiny (15.30\%)}}} &
\rotatebox{90}{\shortstack[l]{\textbf{\textcolor{sidewalk}{\rule{8pt}{8pt}} sidewalk}\\{\tiny (11.13\%)}}} &
\rotatebox{90}{\shortstack[l]{\textbf{\textcolor{parking}{\rule{8pt}{8pt}} parking}\\{\tiny (1.12\%)}}} &
\rotatebox{90}{\shortstack[l]{\textbf{\textcolor{otherground}{\rule{8pt}{8pt}} other-ground}\\{\tiny (0.56\%)}}} &
\rotatebox{90}{\shortstack[l]{\textbf{\textcolor{building}{\rule{8pt}{8pt}} building}\\{\tiny (14.1\%)}}} &
\rotatebox{90}{\shortstack[l]{\textbf{\textcolor{car}{\rule{8pt}{8pt}} car}\\{\tiny (3.92\%)}}} &
\rotatebox{90}{\shortstack[l]{\textbf{\textcolor{truck}{\rule{8pt}{8pt}} truck}\\{\tiny (0.16\%)}}} &
\rotatebox{90}{\shortstack[l]{\textbf{\textcolor{bicycle}{\rule{8pt}{8pt}} bicycle}\\{\tiny (0.03\%)}}} &
\rotatebox{90}{\shortstack[l]{\textbf{\textcolor{motorcycle}{\rule{8pt}{8pt}} motorcycle}\\{\tiny (0.03\%)}}} &
\rotatebox{90}{\shortstack[l]{\textbf{\textcolor{othervehicle}{\rule{8pt}{8pt}} other-veh.}\\{\tiny (0.20\%)}}} &
\rotatebox{90}{\shortstack[l]{\textbf{\textcolor{vegetation}{\rule{8pt}{8pt}} vegetation}\\{\tiny (39.3\%)}}} &
\rotatebox{90}{\shortstack[l]{\textbf{\textcolor{trunk}{\rule{8pt}{8pt}} trunk}\\{\tiny (0.51\%)}}} &
\rotatebox{90}{\shortstack[l]{\textbf{\textcolor{terrain}{\rule{8pt}{8pt}} terrain}\\{\tiny (9.17\%)}}} &
\rotatebox{90}{\shortstack[l]{\textbf{\textcolor{person}{\rule{8pt}{8pt}} person}\\{\tiny (0.07\%)}}} &
\rotatebox{90}{\shortstack[l]{\textbf{\textcolor{bicyclist}{\rule{8pt}{8pt}} bicyclist}\\{\tiny (0.07\%)}}} &
\rotatebox{90}{\shortstack[l]{\textbf{\textcolor{motorcyclist}{\rule{8pt}{8pt}} motorcyclist}\\{\tiny (0.05\%)}}} &
\rotatebox{90}{\shortstack[l]{\textbf{\textcolor{fence}{\rule{8pt}{8pt}} fence}\\{\tiny (3.90\%)}}} & 
\rotatebox{90}{\shortstack[l]{\textbf{\textcolor{pole}{\rule{8pt}{8pt}} pole}\\{\tiny (0.29\%)}}} &
\rotatebox{90}{\shortstack[l]{\textbf{\textcolor{trafficsign}{\rule{8pt}{8pt}} traf.-sign}\\{\tiny (0.08\%)}}}
\\

\midrule
MonoScene\cite{caoMonosceneMonocular3d2022a} & 34.16 & 11.08 & 54.7 & 27.1 & 24.8 & 5.7 & 14.4 & 18.8 & 3.3 & 0.5 & 0.7 & 4.4 & 14.9 & 2.4 & 19.5 & 1.0 & 1.4 & 0.4 & 11.1 & 3.3 & 2.1 \\
TPVFormer$^\dagger$\cite{huangTriperspectiveViewVisionbased2023a} & 34.25 & 11.26 & 55.1 & 27.2 & 27.4 & 6.5 & 14.8 & 19.2 & 3.7 & 1.0 & 0.5 & 2.3 & 13.9 & 2.6 & 20.4 & 1.1 & 2.4 & 0.3 & 11.0 & 2.9 & 1.5 \\
VoxFormer$^\mathbf{T}$\cite{liVoxformerSparseVoxel2023a}c & 43.21 & 13.41 & 54.1 & 26.9 & 25.1 & 7.3 & 23.5 & 21.7 & 3.6 & 1.9 & 1.6 & 4.1 & 24.4 & 8.1 & 24.2 & 1.6 & 1.1 & 0.0 & 13.1 & 6.6 & 5.7 \\
OccFormer$^\mathbf{T}$\cite{liVoxformerSparseVoxel2023a}c & 34.53 & 12.32 & 55.9 & 30.3 & 31.5 & 6.5 & 15.7 & 21.6 & 1.2 & 1.5 & 1.7 & 3.2 & 16.8 & 3.9 & 21.3 & 2.2 & 1.1 & 0.2 & 11.9 & 3.8 & 3.7 \\
NDC-Scene$^\mathbf{T}$\cite{liVoxformerSparseVoxel2023a}c & 33.87 & 11.55 & 56.2 & 28.7 & 28.0 & 5.6 & 15.8 & 19.7 & 1.8 & 1.1 & 1.1 & 4.9 & 14.3 & 2.6 & 20.6 & 0.7 & 1.7 & 0.4 & 11.2 & 3.2 & 1.7 \\
Symphonies$^\mathbf{T}$\cite{liVoxformerSparseVoxel2023a} & 42.19 & 15.04 & 58.4 & 29.3 & 26.9 & 11.7 & 24.7 & 23.6 & 3.2 & 3.6 & 2.6 & 5.6 & 24.2 & 10.0 & 23.1 & \underline{3.2} & 1.9 & \underline{2.0} & 16.1 & 7.7 & 8.0 \\
MVFormer$^\mathbf{T}$\cite{meiCamerabased3dsemantic2024a} & 43.02 & 15.81 & 59.7 & 31.5 & 29.5 & \underline{12.5} & 24.4 & 23.9 & 2.8 & 3.7 & 2.4 & 6.2 & 26.8 & 11.2 & 24.9 & 3.1 & 2.6 & \textbf{2.2} & 16.8 & 8.4 & 
7.8 \\
CGFormer\cite{yuContextgeometryaware2024} & 44.41 & 16.63 & 64.3 & 34.2 & 34.1 & 12.1 & 25.8 & 26.1 & 4.3 & 3.7 & 1.3 & 2.7 & 24.5 & 11.2 & \underline{29.3} & 1.7 & \underline{3.6} & 0.4 & 18.7 & 8.7 & \underline{9.3} \\
HTCL$^\mathbf{T}$\cite{liHierarchicalTemporalContext2025a} & 44.23 & 17.09 & \underline{64.4} & \underline{34.8} & 33.8 & 12.4 & 25.9 & \underline{27.3} & 5.7 & 1.8 & 2.2 & 5.4 & 25.3 & 10.8 & \textbf{31.2} & 1.1 & 
3.1 & 0.9 & \textbf{21.1} & 9.0 & 8.3 \\
DISC\cite{liu2025disentangling} & \underline{45.32} & 17.35 & 63.1 & 34.7 & \underline{34.6} & \textbf{12.6} & 26.6 & 26.7 & 5.5 & \underline{4.0} & \underline{4.7} & \textbf{8.1} & 26.5 & 10.3 & \underline{29.3} & 2.8 & 2.5 & 1.1 & 19.3 & 8.4 & 8.7 \\
VisHall3D\cite{lu2025vishall3d} & \textbf{46.5} & \underline{17.46} & \textbf{64.6} & 34.1 & 32.0 & \underline{12.5} & \underline{26.9} & 26.7 & \underline{7.5} & 2.9 & 3.3 & 6.2 & \textbf{27.3} & \underline{12.5} & 28.0 & 2.3 & \textbf{5.1} & 1.9 & 19.5 & \underline{9.2} & 9.2 \\
VOIC(proposed) & 45.22 & \textbf{18.01} & 63.3 & \textbf{36.0} & \textbf{36.9} & 8.2 & \textbf{27.5} & \textbf{27.9} & \textbf{8.3} & \textbf{5.9} & \textbf{6.0} & \underline{7.0} & \underline{27.2} & \textbf{12.6} 
& 28.6 & \textbf{3.6} & 2.4 & 0.2 & \underline{20.9} & \textbf{9.8} & \textbf{10.0} \\
\bottomrule
\end{tabular}%
}

\end{table*}

\begin{table*}[ht]
\caption{Quantitative results on the  SSCBench-KITTI-360 test set. $^\mathbf{T}$ represents the use of multi-frame input. $^\dagger$ represents the use of multi-camera images as input. The best and the second best results are marked in \textbf{bold} and $\underline{\text{underlined}}$, respectively.}
\label{tab:kitti360_results}
\centering
\small
\resizebox{\linewidth}{!}{%
\begin{tabular}{l|cc|*{18}{c}}
\toprule
\textbf{Method} & \textbf{IoU}& \textbf{mIoU} &
\rotatebox{90}{\shortstack[l]{\textbf{\textcolor{blue!80!black}{\rule{8pt}{8pt}} car}\\{\tiny (2.85\%)}}} &
\rotatebox{90}{\shortstack[l]{\textbf{\textcolor{cyan!50}{\rule{8pt}{8pt}} bicycle}\\{\tiny (0.01\%)}}} &
\rotatebox{90}{\shortstack[l]{\textbf{\textcolor{blue!100}{\rule{8pt}{8pt}} motorcycle}\\{\tiny (0.00\%)}}} &
\rotatebox{90}{\shortstack[l]{\textbf{\textcolor{violet!40}{\rule{8pt}{8pt}} truck}\\{\tiny (0.16\%)}}} &
\rotatebox{90}{\shortstack[l]{\textbf{\textcolor{green!70!black}{\rule{8pt}{8pt}} other-veh.}\\{\tiny (0.16\%)}}} &
\rotatebox{90}{\shortstack[l]{\textbf{\textcolor{red!100}{\rule{8pt}{8pt}} person}\\{\tiny (0.02\%)}}} &
\rotatebox{90}{\shortstack[l]{\textbf{\textcolor{magenta!100}{\rule{8pt}{8pt}} road}\\{\tiny (14.96\%)}}} &
\rotatebox{90}{\shortstack[l]{\textbf{\textcolor{pink!70}{\rule{8pt}{8pt}} parking}\\{\tiny (2.51\%)}}} &
\rotatebox{90}{\shortstack[l]{\textbf{\textcolor{purple}{\rule{8pt}{8pt}} sidewalk}\\{\tiny (6.43\%)}}} &
\rotatebox{90}{\shortstack[l]{\textbf{\textcolor{red!70!black}{\rule{8pt}{8pt}} other-grnd.}\\{\tiny (0.64\%)}}} &
\rotatebox{90}{\shortstack[l]{\textbf{\textcolor{yellow}{\rule{8pt}{8pt}} building}\\{\tiny (15.67\%)}}} &
\rotatebox{90}{\shortstack[l]{\textbf{\textcolor{orange}{\rule{8pt}{8pt}} fence}\\{\tiny (0.96\%)}}} &
\rotatebox{90}{\shortstack[l]{\textbf{\textcolor{green!100!black}{\rule{8pt}{8pt}} vegetation}\\{\tiny (41.99\%)}}} &
\rotatebox{90}{\shortstack[l]{\textbf{\textcolor{green!50!yellow}{\rule{8pt}{8pt}} terrain}\\{\tiny (7.10\%)}}} &
\rotatebox{90}{\shortstack[l]{\textbf{\textcolor{yellow!50!brown}{\rule{8pt}{8pt}} pole}\\{\tiny (0.22\%)}}} &
\rotatebox{90}{\shortstack[l]{\textbf{\textcolor{red!90!black}{\rule{8pt}{8pt}} traf.-sign}\\{\tiny (0.52\%)}}} &
\rotatebox{90}{\shortstack[l]{\textbf{\textcolor{cyan!70!red}{\rule{8pt}{8pt}} other-struct.}\\{\tiny (0.28\%)}}} &
\rotatebox{90}{\shortstack[l]{\textbf{\textcolor{cyan!100}{\rule{8pt}{8pt}} other-obj.}\\{\tiny (0.58\%)}}} \\
\midrule
MonoScene\cite{caoMonosceneMonocular3d2022a} & 37.87 & 12.31 & 19.3 & 0.4 & 0.6 & 8.0 & 2.0 & 0.9 & 48.4 & 11.4 & 28.1 & 3.2 & 32.9 & 3.5 & 26.2 & 16.8 & 6.9 & 5.7 & 4.2 & 3.1 \\
TPVFormer$^\dagger$\cite{huangTriperspectiveViewVisionbased2023a} & 40.22 & 13.64 & 21.6 & 1.1 & 1.4 & 8.1 & 2.6 & 2.4 & 53.0 & 12.0 & 31.1 & 3.8 & 34.8 & 4.8 & 30.1 & 17.5 & 7.5 & 5.9 & 5.5 & 2.7 \\
OccFormer\cite{zhangOccformerDualpathTransformer2023a} & 40.27 & 13.81 & 22.6 & 0.7 & 0.3 & 9.9 & 3.8 & 2.8 & 54.3 & 13.4 & 31.5 & 3.6 & 36.4 & 4.8 & 31.0 & 19.5 & 7.8 & 8.5 & 7.0 & 4.6 \\
VoxFormer$^\mathbf{T}$\cite{liVoxformerSparseVoxel2023a} & 38.76 & 11.91 & 17.8 & 1.2 & 0.9 & 4.6 & 2.1 & 1.6 & 47.0 & 9.7 & 27.2 & 2.9 & 31.2 & 5.0 & 29.0 & 14.7 & 6.5 & 6.9 & 3.8 & 2.4 \\
CGFormer\cite{yuContextgeometryaware2024} & 48.07 & 20.05 & 29.9 & 3.4 & 4.0 & 17.6 & 6.8 & 6.6 & \underline{63.9} & 17.2 & 40.7 & \underline{5.5} & 42.7 & 8.2 & 38.8 & \textbf{24.9} & 16.2 & 17.5 & 10.2 & 6.8 \\   
DISC\cite{liu2025disentangling}  & 47.35 & 20.55 & 29.4 & \textbf{4.6} & \textbf{8.2} & \underline{19.2} & 8.5 & 6.7 & 61.9 & \underline{17.5} & 40.1 & 5.3 & 42.5 & \underline{9.2} & 38.7 & 23.0 & \underline{16.7} & 19.5 & \underline{10.3} & \underline{8.2} \\
VisHall3D\cite{lu2025vishall3d} & \textbf{49.12} & \underline{20.95} & \underline{30.8} & 1.9 & 6.6 & 18.0 & \underline{8.7} & \textbf{8.7} & \textbf{64.4} & \textbf{18.8} & \textbf{41.5} & 4.5 & \textbf{43.9} & 9.1 & \textbf{39.8} & \underline{24.9} & 16.5 & \textbf{20.7} & 10.3 & 8.0 \\
VOIC(Ours) & \underline{48.18} & \textbf{21.37} & \textbf{30.9} & \underline{4.4} & \underline{7.2} & \textbf{21.0} & \textbf{9.8} & \underline{8.4} & 62.6 & 11.9 & \underline{41.3} & \textbf{6.1} & \underline{43.5} & \textbf{11.0} & \underline{38.9} & 22.7 & \textbf{17.2} & \underline{20.1} & \textbf{11.9} & \textbf{9.2} \\
\bottomrule
\end{tabular}%
}

\end{table*}

\begin{table}[!t]
\caption{Comparison of inference time and number of parameters}
\label{tab:Comparison}
\centering
\begin{tabular}{l | c c c}
\toprule
\textbf{Method} & \textbf{mIoU (\%)$\uparrow$} & \textbf{Times (s)$\downarrow$} & \textbf{Params (M)$\downarrow$} \\
\midrule
MonoScene\cite{caoMonosceneMonocular3d2022a} & 11.08 & 0.274 & 132.4 \\
OccFormer\cite{zhangOccformerDualpathTransformer2023a} & 13.46 & 0.338 & 203.4 \\
VoxFormer\cite{liVoxformerSparseVoxel2023a} & 13.35 & 0.256 & 57.9 \\
Symphonies\cite{jiangSymphonize3dSemantic2024} & 14.89 & 0.319 & 59.3 \\
VisHall3D\cite{lu2025vishall3d} & 17.46 & 0.340 & 127.8 \\
VOIC(Ours) & \textbf{18.01} & \textbf{0.243} & \textbf{45.4} \\
\bottomrule
\end{tabular}

\end{table}

\textbf{Quantitative Results.} Table~\ref{tab:semantickitti_results} presents a comparison of our proposed VOIC with other state-of-the-art camera-based SSC methods on the SemanticKITTI hidden test set. VOIC achieves the best mIoU among the compared methods on the SemanticKITTI hidden test set, while maintaining competitive IoU performance. Compared with CGFormer under the identical single-view RGB and predicted-depth input setting, our method improves mIoU by 1.38\% and IoU by 0.81\%. These results validate the effectiveness of the proposed design in semantic scene completion while maintaining competitive geometric completion performance. Notably, VOIC delivers top-tier performance on long-tailed categories, such as car, truck, bicycle, and motorcycle.

To further evaluate the effectiveness of VOIC across diverse scenarios, as shown in Table~\ref{tab:kitti360_results}, our method demonstrates significant advantages in semantic and geometric analysis on the data-rich SSCBench-KITTI-360 benchmark. VOIC surpasses all published methods in mIoU metrics, achieving the best performance across the majority of categories.

Furthermore, as detailed in Table~\ref{tab:Comparison}, we compare the inference time and number of parameters with other state-of-the-art methods on the SemanticKITTI validation set. While achieving strong semantic performance with 18.01\% mIoU, VOIC also features a shorter inference time and a more lightweight architecture, requiring only 45.4M parameters.

\begin{figure*}[t]
    \centering
    \includegraphics[width=\textwidth]{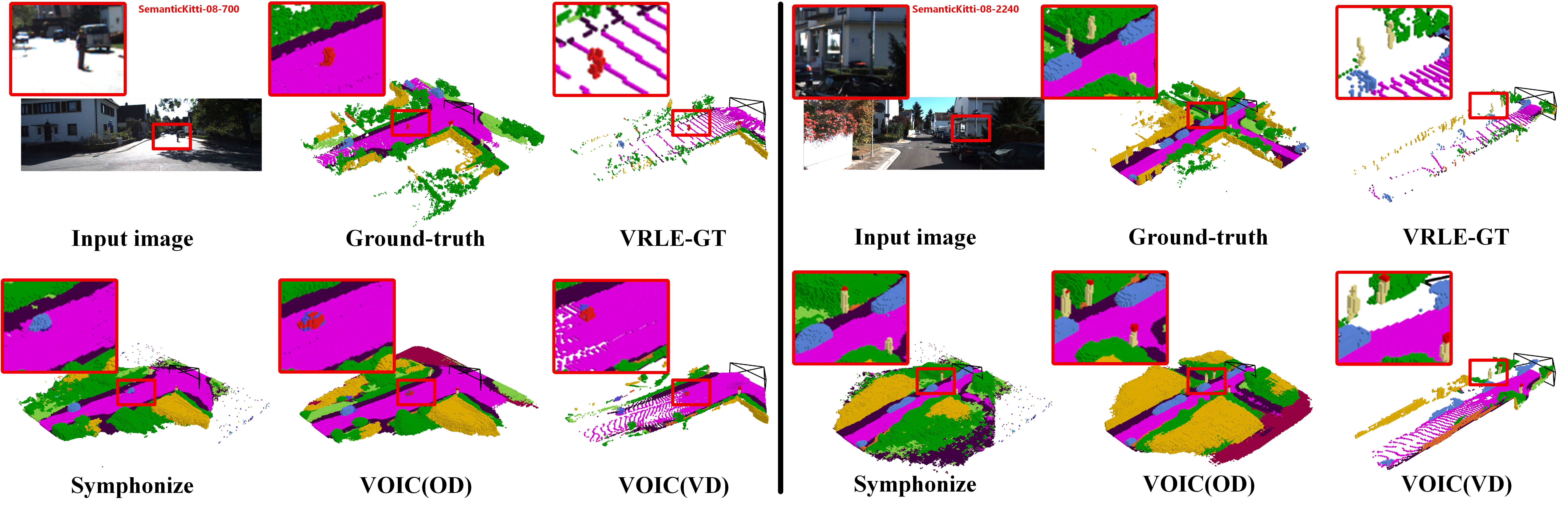}

    \begin{minipage}{0.95\textwidth}
    \centering
    \scriptsize
    \textcolor{car}{\rule{6pt}{6pt}}\,car\hspace{0.7em}%
    \textcolor{bicycle}{\rule{6pt}{6pt}}\,bicycle\hspace{0.7em}%
    \textcolor{motorcycle}{\rule{6pt}{6pt}}\,motorcycle\hspace{0.7em}%
    \textcolor{truck}{\rule{6pt}{6pt}}\,truck\hspace{0.7em}%
    \textcolor{othervehicle}{\rule{6pt}{6pt}}\,other-vehicle\hspace{0.7em}%
    \textcolor{person}{\rule{6pt}{6pt}}\,person\hspace{0.7em}%
    \textcolor{bicyclist}{\rule{6pt}{6pt}}\,bicyclist\hspace{0.7em}%
    \textcolor{motorcyclist}{\rule{6pt}{6pt}}\,motorcyclist\hspace{0.7em}%
    \textcolor{road}{\rule{6pt}{6pt}}\,road\hspace{0.7em}%
    \textcolor{parking}{\rule{6pt}{6pt}}\,parking\hspace{0.7em}%
    \textcolor{sidewalk}{\rule{6pt}{6pt}}\,sidewalk\hspace{0.7em}%
    \textcolor{otherground}{\rule{6pt}{6pt}}\,other-ground\hspace{0.7em}%
    \textcolor{building}{\rule{6pt}{6pt}}\,building\hspace{0.7em}%
    \textcolor{fence}{\rule{6pt}{6pt}}\,fence\hspace{0.7em}%
    \textcolor{vegetation}{\rule{6pt}{6pt}}\,vegetation\hspace{0.7em}%
    \textcolor{trunk}{\rule{6pt}{6pt}}\,trunk\hspace{0.7em}%
    \textcolor{terrain}{\rule{6pt}{6pt}}\,terrain\hspace{0.7em}%
    \textcolor{pole}{\rule{6pt}{6pt}}\,pole\hspace{0.7em}%
    \textcolor{trafficsign}{\rule{6pt}{6pt}}\,traffic-sign
    \end{minipage}
    \caption{
    Qualitative results on the SemanticKITTI validation set. 
    VOIC enhances overall scene classification through high-quality visible-range semantic priors predicted by VD. 
    Shown here is a qualitative comparison between Symphonies\cite{jiangSymphonize3dSemantic2024} and VOIC on representative validation scenes.
    }
    \label{fig:visual_result}

\end{figure*}

\textbf{Qualitative Results.} To provide an intuitive demonstration of VOIC's performance, Figure~\ref{fig:visual_result} illustrates qualitative results on the SemanticKITTI validation set, comparing our method with Symphonies. The first row shows the input images, ground-truth labels, and VRLE-GT, while the second row presents the outputs of Symphonies and the OD/VD outputs of VOIC. It can be observed that VD closely follows VRLE-GT, achieving more accurate semantic segmentation and occupancy information on visible objects. For example, in the first case, VOIC effectively captures the position and details of small objects such as cars, while in the second case, it accurately predicts pedestrian locations and guides OD accordingly.

\subsection{Ablation Studies}
\label{sec:ablation}

\textbf{Analysis of VRLE Supervision on Visible Regions.}
We further analyze the effect of different supervision strategies using VRLE-mIoU, which evaluates semantic accuracy only on voxels marked as visible by the VRLE mask. To provide a complete view of scene understanding, we also report the overall mIoU computed over the full scene.

As shown in Table~\ref{tab:vrle_miou}, VOIC achieves the best performance among all compared methods, with 30.78\% VRLE-mIoU and 18.00\% overall mIoU. Compared with MonoScene, Symphonies, and VisHall3D, the proposed method improves both visible-region semantic perception and holistic scene completion. This indicates that strengthening visible-region learning can provide more reliable priors for complete-scene semantic reasoning.

The internal variants further reveal the importance of the supervision design. When only VRLE supervision is used, the model obtains 22.78\% VRLE-mIoU but only 6.05\% overall mIoU. This result shows that visible-region supervision alone cannot support complete scene completion, since the model lacks sufficient guidance for occluded and unobserved regions. In contrast, directly summing the visible and global labels improves VRLE-mIoU to 28.31\%, but the overall mIoU remains limited at 10.00\%. This suggests that simply merging different supervision targets does not effectively resolve the learning conflict between visible-region perception and holistic completion.

Removing VRLE from the full framework leads to 27.93\% VRLE-mIoU and 16.05\% overall mIoU. Since this variant preserves the main architecture but removes explicit visibility-aware supervision, the performance gap demonstrates that the improvement of VOIC is not merely caused by the dual-decoder structure, but also relies on the proposed VRLE-based supervision strategy.

Overall, these results show that effective SSC requires both reliable visible-region supervision and global scene-level learning. Visible-only supervision is insufficient for holistic completion, while naive label summation weakens the distinction between visible and occluded learning objectives. By explicitly decoupling these supervision signals within a dual-decoder framework, VOIC achieves better visible-region perception and stronger complete-scene semantic modeling.
\begin{table}[!t]
\centering
\caption{Comparison of VRLE-mIoU and overall mIoU on the SemanticKITTI validation set.}
\label{tab:vrle_miou}
\begin{tabular}{lcc}
\toprule
\textbf{Method} & \textbf{VRLE-mIoU (\%)} & \textbf{mIoU (\%)} \\
\midrule
MonoScene~\cite{caoMonosceneMonocular3d2022a} & 19.86 & 11.30 \\
Symphonies~\cite{jiangSymphonize3dSemantic2024} & 27.37 & 14.89 \\
VisHall3D~\cite{lu2025vishall3d} & 29.08 & 16.31 \\
\midrule
VOIC (VRLE-only) & 22.78 & 6.05 \\
VOIC (w/o VRLE) & 27.93 & 16.05 \\
VOIC (label summation) & 28.31 & 10.00 \\
VOIC (full) & \textbf{30.78} & \textbf{18.00} \\
\bottomrule
\end{tabular}
\end{table}

\begin{table}[!t]
\caption{Ablation Study on Architecture, VRLE, and 2D-to-3D Projection}
\label{tab:ablation}
\centering
\resizebox{\columnwidth}{!}{%
\begin{tabular}{c|ccc|cc}
\toprule
\textbf{Method} & \textbf{VRLE} & \textbf{\makecell{2D-to-3D \\ Projection}} & \textbf{Decoder} & \textbf{IoU (\%)} & \textbf{mIoU (\%)} \\
\midrule
Baseline & \ding{55} & FLoSP & OD & 36.86 & 11.09 \\
1 & \ding{55} & VEFC & OD & 44.88 & 15.88 \\
2 & \ding{55} & VEFC & VD+OD & 44.93 & 16.09 \\
VOIC(Ours) & \ding{51} & VEFC & VD+OD & 45.53 & 18.06 \\
\bottomrule
\end{tabular}%
}

\end{table}

To better understand the contributions of different components in VOIC, we conducted a series of ablation experiments, focusing on three aspects: the architectural design, the choice of 2D-to-3D projection methods, and the interaction flow between the Visible Decoder (VD) and the Occlusion Decoder (OD). All results are benchmarked on the SemanticKITTI validation set.

\textbf{Ablation on the Architectural Components and Projection Method.}
Table~\ref{tab:ablation} summarizes the performance of VOIC under different architectural configurations. We first construct a Baseline model by removing the $\text{VRLE}$ label, adopting $\text{FLoSP}$\cite{caoMonosceneMonocular3d2022a} as the 2D-to-3D projection module, and using $\text{OD}$ as the final decoder, which yields $36.86\%$ $\text{IoU}$ and $11.09\%$ $\text{mIoU}$.

Replacing $\text{FLoSP}$ with the proposed $\text{VEFC}$ module (Method 1), while still omitting $\text{VRLE}$ and using only $\text{OD}$, improves performance to $44.88\%$ $\text{IoU}$ and $15.88\%$ $\text{mIoU}$. This improvement arises from $\text{VEFC}$'s depth-aware feature aggregation, which provides more consistent geometric cues for semantic lifting.

Introducing the dual-decoder cascade $\text{VD+OD}$ (Method 2), where $\text{VEFC}$ is used but $\text{VRLE}$ is still omitted, further improves performance to $44.93\%$ $\text{IoU}$ and $16.09\%$ $\text{mIoU}$. In this setting, VD does not receive explicit visible-region supervision; instead, it is optimized indirectly through gradients backpropagated from the OD loss.

Finally, incorporating the $\text{VRLE}$ label as explicit supervision for $\text{VD}$ (VOIC, Ours) achieves the best results of {$45.53\%$ $\text{IoU}$ and $18.06\%$ $\text{mIoU}$.} This progression demonstrates the crucial role of $\text{VRLE}$ in enabling VD to progressively refine visible-region representations and produce a clear separation between visible and occluded region reconstruction.

\begin{table}[!t]
\caption{Interaction Flow Ablation}
\label{tab:VD-OD}
\centering
\begin{tabular}{c|c|cc}
\toprule
Method & Interaction Flow & IoU (\%) & mIoU (\%) \\
\midrule
1 & None & 44.85 & 15.56 \\
2 & VD $\rightarrow$ OD & 45.33 & 17.02 \\
VOIC(Ours)&VD $\leftrightarrow$ OD & {45.53} & {18.06} \\
\bottomrule
\end{tabular}

\end{table}

\textbf{Ablation on the Decoder Interaction Flow.}
Table~\ref{tab:VD-OD} summarizes the effect of information exchange between $\text{VD}$ and $\text{OD}$. When the two decoders are isolated, the model obtains $15.56\%$ mIoU. Adding a unidirectional link from VD to OD raises the performance to $17.02\%$ mIoU, indicating that visible-region features provide useful priors but a single forward fusion remains limited.
Since VD guides OD through visible-region predictions, it naturally becomes necessary to propagate OD’s global information back to VD to strengthen VD’s feature learning, a capability absent in the unidirectional design.
We therefore introduce a reverse link from OD to VD. This feedback brings OD’s global context into VD, enabling mutual refinement and improving the final performance to {$18.06\%$ mIoU}.
Such bidirectional interaction allows both decoders to share complementary cues, benefiting full-scene reconstruction and reducing inconsistencies in the visible regions.

\section{Conclusion}
In this paper, we introduced the Visible--Occluded Interactive Completion Network (VOIC), a problem-driven dual-decoder framework for single-view RGB-based SSC with external depth guidance. Our approach is grounded in the observation that traditional uniform global supervision fails to account for the reliability differences between visible and occluded regions, leading to entangled optimization and suboptimal feature learning. By decoupling the task into visible-region perception and holistic completion, VOIC effectively addresses this optimization mismatch and fosters a synergistic relationship between observed and unobserved spaces.

The core of our method lies in the Visible Region Label Extraction (VRLE) strategy, which provides the Visible Decoder (VD) with clean, visibility-aware objectives to establish reliable geometric and semantic priors. These priors act as a structured foundation for the Occlusion Decoder (OD), which fills in the occluded geometry under global annotations. Furthermore, the bidirectional interactive reasoning between the two decoders ensures that visible features benefit from global contextual feedback while maintaining spatial consistency across the entire scene.

Extensive evaluations on the SemanticKITTI and SSCBench-KITTI-360 benchmarks demonstrate that VOIC achieves competitive performance, particularly in resolving ambiguities in complex occluded regions. By addressing the intrinsic observability differences through targeted supervision and interactive decoding, VOIC provides a robust and effective solution for camera-based 3D scene understanding.

\bibliographystyle{IEEEtran}
\bibliography{VOIC}

\end{document}